\documentclass[pmlr,twocolumn,10pt]{jmlr} 

\mlhtrack{proceedings}

\newif\iffinal
\finaltrue  

\iffinal
    \ifmlhneedspmlr
      \jmlrvolume{346}
      \jmlryear{2026}
    \fi
    \ifmlhfindings \jmlrproceedings{}{ML4H 2026 - Findings Track}\fi
    \ifmlhdemo     \jmlrproceedings{}{ML4H 2026 - Demo Track}\fi
    \jmlrworkshop{Machine Learning for Health (ML4H) 2026}
\else
    \jmlrproceedings{}{Submitted to ML4H 2026: \mlhtrackname}
    \jmlrworkshop{Machine Learning for Health (ML4H) 2026}
\fi

\usepackage{array}
\usepackage{booktabs}
\usepackage{siunitx}

\usepackage[switch]{lineno}

\usepackage{adjustbox}

\theorembodyfont{\upshape}
\theoremheaderfont{\scshape}
\theorempostheader{:}
\theoremsep{\newline}

\newcommand{\benchmark}{CXReasonDial\xspace}
\newcommand{\agent}{CXReasonAgent\xspace}

\definecolor{skyblue}{RGB}{70,160,220}

\title[\agent: Evidence-Grounded Diagnostic Reasoning Agent for Chest X-rays]{\agent: Evidence-Grounded \\ Diagnostic Reasoning Agent for Chest X-rays
\\[0.4em]
{\normalsize
\raisebox{-0.35em}{\includegraphics[height=1.5em]{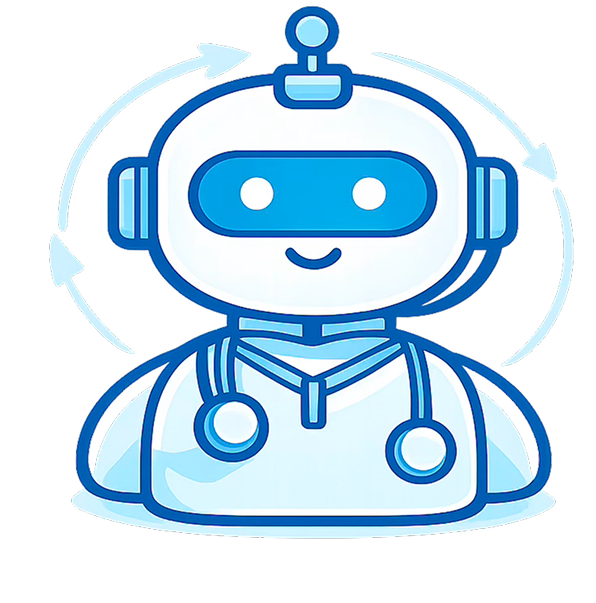}}
\ \href{https://ttumyche.github.io/cxreasonagent/}{\textcolor{skyblue}{Project Page}}
}
}

\author{%
\Name{Hyungyung Lee\(^{\dagger}\)}\thanks{Work done while at KAIST.}\Email{Hyungyung\_Lee@a-star.edu.sg}\\
\Name{Hangyul Yoon\(^{\ddagger}\)} \Email{hangyulmd@kaist.ac.kr}\\
\Name{Edward Choi\(^{\ddagger}\)} \Email{edwardchoi@kaist.ac.kr}\\
\addr \(^{\dagger}\)Agency for Science, Technology and Research (A*STAR), Singapore\\
\addr \(^{\ddagger}\)Korea Advanced Institute of Science \& Technology, South Korea
}

\begin{document}

\maketitle

\ifmlhdemo\else
\begin{abstract}
Chest X-ray plays a central role in thoracic diagnosis, and its interpretation inherently requires multi-step, evidence-grounded reasoning.
However, large vision-language models (LVLMs) often generate plausible responses that are not faithfully grounded in diagnostic evidence and provide limited visual evidence for verification, while also requiring costly retraining to support new diagnostic tasks, limiting their reliability and adaptability in clinical settings.
To address these limitations, we present \agent, a diagnostic agent that integrates a large language model (LLM) with clinically grounded diagnostic tools to perform evidence-grounded diagnostic reasoning using image-derived diagnostic and visual evidence.
To evaluate these capabilities, we introduce \benchmark, a multi-turn dialogue benchmark with 1,946 dialogues across 12 diagnostic tasks, and show that \agent produces more faithfully grounded responses than LVLMs. These findings highlight the importance of integrating clinically grounded diagnostic tools to enable reliable and verifiable diagnostic reasoning.
\end{abstract}

\fi

\ifmlhneedsstatements
\paragraph*{Data and Code Availability}
This study uses MIMIC-CXR-JPG \citep{johnson2019mimic}, which is available to credentialed users through PhysioNet under a data use agreement.
We plan to release CXReasonDial through PhysioNet, subject to the applicable data use agreement and PhysioNet's review and approval.
An interactive demo of CXReasonAgent is available at \href{https://ttumyche.github.io/CXReasonAgent_Demo/}{Link}.

\paragraph*{Institutional Review Board (IRB)}
This research does not require IRB approval.
\fi

\section{Introduction}
\label{sec:intro}

\begin{figure*}[htb!]
\includegraphics[width=\textwidth]{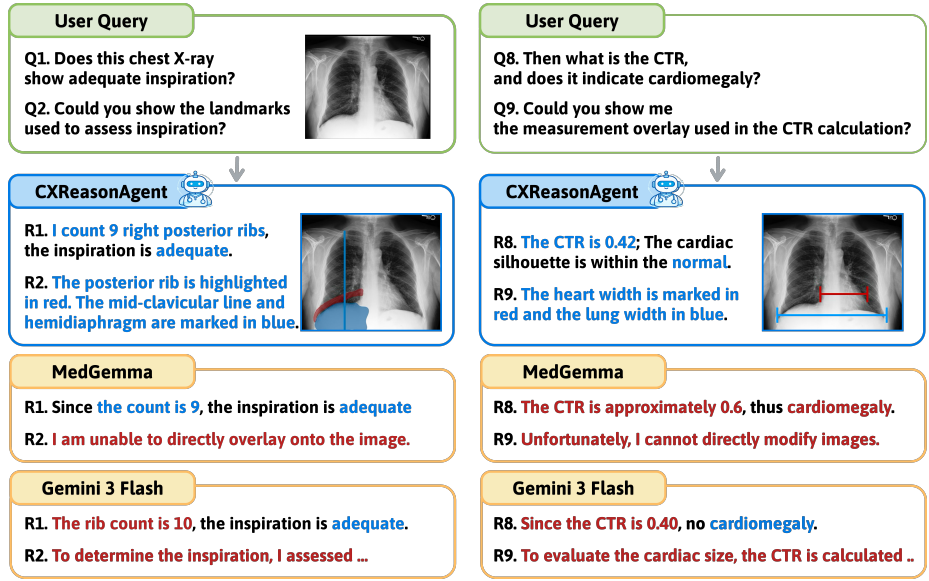}
\caption{\textbf{Qualitative examples on \benchmark}.
\agent produces responses faithfully grounded in image-derived diagnostic and visual evidence, whereas LVLMs often generate ungrounded responses and fail to provide visual evidence.
}
\label{fig:example}
\end{figure*}

Chest X-ray (CXR) is a primary imaging modality for thoracic diagnosis and is widely used in clinical practice \citep{bansal2019interpreting}.
Inherently, CXR interpretation is a multi-step, evidence-grounded reasoning process that involves identifying relevant anatomical regions, deriving quantitative measurements or spatial observations, and applying diagnostic criteria \citep{delrue2010difficulties}.
Consequently, for diagnostic assistants to be trustworthy and clinically useful, their intermediate diagnostic steps should be grounded in clinically valid and verifiable image-derived diagnostic evidence, and their reasoning must remain coherent and consistent throughout the process.

However, recent studies \citep{NEURIPS2025_7ef1f85b,meddeb2025evaluating,nguyen2025localizing} demonstrate that large vision-language models (LVLMs) often generate plausible responses but are not faithfully grounded in the diagnostic evidence present in the image, thereby undermining the reliability of their conclusions in clinical practice.
Moreover, LVLMs typically present their reasoning through textual explanations alone, which makes it difficult to verify how conclusions are derived from the image.
Effective diagnostic support instead requires presenting diagnostic evidence directly on the image, enabling direct verification.
However, such visual evidence is rarely provided in practice, thereby limiting verifiability of diagnostic reasoning \citep{amann2020explainability,tjoa2020survey}.
Furthermore, building LVLMs that handle diverse diagnostic tasks often introduces significant complexity and inefficiency, motivating tool-augmented diagnostic agents \citep{chen2025radfabric,fallahpour2025medrax,li2024mmedagent} that integrate task-specific models as tools \citep{cohen2022torchxrayvision,li2023llava,wang2023chatcad,zhang2023biomedclip}, enabling extension to new tasks without costly retraining.
However, these approaches typically rely on tools that provide final diagnostic conclusions or region-level visualizations, without exposing the intermediate diagnostic steps used to derive these conclusions from image-derived evidence.
Consequently, they still fall short of supporting reliable and verifiable evidence-grounded diagnostic reasoning.
Notably, recent work in ECG analysis \citep{chung2026ecg} demonstrates that integrating measurement-based tools \citep{makowski2021neurokit2} can provide quantitative diagnostic evidence, pointing toward more reliable diagnostic reasoning.

Based on these observations, we propose \agent, a tool-augmented diagnostic agent that combines an LLM with clinically grounded diagnostic tools to provide verifiable, evidence-grounded diagnostic reasoning through multi-turn interaction. The LLM does not directly access the chest X-ray. Instead, it interprets each user query and the dialogue context to identify the diagnostic evidence required to answer the request and invokes the appropriate tools to retrieve image-derived measurements or spatial observations. It then generates a response that links this evidence and its corresponding clinical criteria to a diagnostic conclusion and, when requested, returns a visual annotation that presents the relevant evidence directly on the chest X-ray.
To evaluate these abilities, we introduce \benchmark, a multi-turn dialogue benchmark comprising 1,946 dialogues spanning 12 diagnostic tasks, designed to measure whether responses are correctly grounded in image-derived evidence, reflecting the iterative nature of diagnostic reasoning in user-assistant interactions.
Experimental results show that \agent produces correctly grounded responses, whereas LVLMs often generate ungrounded responses (Fig.~\ref{fig:example}).
These findings highlight the importance of integrating clinically grounded diagnostic tools for reliable and verifiable evidence-grounded diagnostic 
reasoning.

\section{Related Works}
\noindent
\textbf{Medical LVLMs and Evidence Grounding}
Medical LVLMs enable conversational assistance for medical images, including chest X-rays, by combining image understanding with natural-language interaction \citep{li2023llava,thirunavukarasu2023large, sellergren2025medgemma}.
However, recent studies have reported limitations in their diagnostic reliability and visual grounding \citep{NEURIPS2025_7ef1f85b, meddeb2025evaluating, nguyen2025localizing}, as well as sensitivity to image artefacts \citep{cheng2025understanding}.
Models may also generate quantitative statements that are not supported by the image \citep{heiman2025factchexcker}.
These findings show that LVLMs may generate clinically plausible responses without being reliably grounded in diagnostic evidence from the image, making it difficult to verify how their diagnostic claims are supported.
To address these limitations, \agent differs from LVLMs by using specialized diagnostic tools to retrieve clinically grounded, image-derived evidence and generate responses based on that evidence.

\noindent
\textbf{Tool-Augmented Medical Agents}
Tool augmentation enables language models to select and invoke specialized tools for medical image analysis without retraining a monolithic model.
General multimodal medical agents have demonstrated this capability across medical imaging applications \citep{wang2023chatcad,li2024mmedagent}.
More recently, radiology-focused systems such as RadFabric \citep{chen2025radfabric} and MedRax \citep{fallahpour2025medrax} have integrated multiple specialized chest X-ray analysis tools with language models to address diverse radiology queries.
A key difference is how a diagnostic conclusion is substantiated in the user-facing output.
Prior tool-augmented systems typically provide a predicted finding, a localized region, or a generated report, without explicitly exposing the image-derived evidence and clinical criteria used to derive the conclusion.
In contrast, \agent makes the diagnostic derivation explicit by presenting the image-derived evidence, including the relevant measurement or spatial observation, the clinical criterion applied to that evidence, and the resulting conclusion.
When visual evidence is requested, it presents the relevant measurement or anatomical observation directly on the chest X-ray, rather than only highlighting the predicted finding.

\noindent
\textbf{Medical QA and Dialogue Benchmarks}
Medical dialogue datasets, such as MedDialog \citep{he2020meddialog}, support multi-turn conversational modeling using text-based patient--doctor conversations. Separately, medical visual question-answering benchmarks evaluate image-based diagnostic questions across diverse medical imaging
domains \citep{lau2018dataset,liu2021slake,he2020pathvqa,nguyen2025localizing}.
More recently, CXReasonBench \citep{NEURIPS2025_7ef1f85b} introduced a chest X-ray benchmark for evaluating multi-stage structured diagnostic reasoning.
However, these medical image QA benchmarks evaluate independent question--answer instances rather than multi-turn user--assistant dialogues. 
ToolDial \citep{shim2025tooldial} introduced a general framework for constructing multi-turn tool-use dialogues through turn-level action sequences. 
We adopt this action-sequence formulation to construct the dialogue scenarios in \benchmark. 
Building on these directions, \benchmark provides multi-turn chest X-ray dialogues across 12 clinically grounded diagnostic tasks.
It evaluates not only whether a response addresses the user query, but also whether it remains grounded in the diagnostic evidence specified for each turn throughout the dialogue.

\section{\agent}
\begin{figure*}[htb!]
\includegraphics[width=\textwidth]{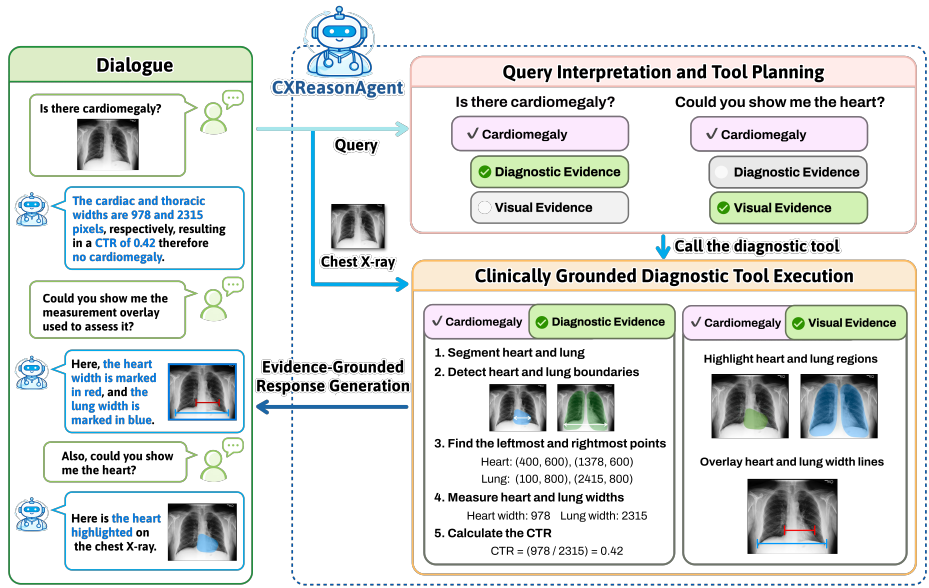}
\caption{
\textbf{Overview of \agent}.
The pipeline comprises three stages: (1) interpreting the user query and planning the appropriate diagnostic tool call, (2) constructing diagnostic and visual evidence from the chest X-ray using clinically grounded diagnostic 
tools, and (3) generating responses solely grounded in this evidence.
}
\label{fig:agent}
\end{figure*}

\noindent
As shown in Fig.~\ref{fig:agent}, \agent is a tool-augmented diagnostic agent that combines a large language model (LLM) with clinically grounded diagnostic tools to provide evidence-grounded diagnostic reasoning through multi-turn interaction.
The LLM does not directly access the chest X-ray. 
Instead, it uses the user query and dialogue history to identify the diagnostic evidence needed to answer the request and select the appropriate tool call.
The selected tool deterministically extracts image-derived measurements or spatial observations from the chest X-ray according to predefined, clinically grounded assessment procedures.
It then applies the corresponding clinical criteria to this evidence and returns the evidence together with the diagnostic conclusion.
The LLM then generates a response that explicitly links the returned image-derived evidence and its corresponding clinical criterion to the diagnostic conclusion.
To ensure clinical verifiability, the agent operates within 12 predefined diagnostic tasks whose image-derived evidence can be reliably extracted by the integrated tools \citep{NEURIPS2025_7ef1f85b}.
For detailed descriptions of the supported diagnostic tasks, refer to Appendix~\ref{apd:diagnostic_tasks}. The prompts used by the LLM are provided in Appendix~\ref{apd:agent_prompts}.

\noindent
\subsection{Query Interpretation and Tool Planning.}
Given a user query and the dialogue history associated with a chest X-ray, the LLM identifies the diagnostic tasks and the type of evidence required to
answer the request.
Queries are categorized into two types:
\textbf{\textit{1) Diagnostic Evidence Request}}, which asks for image-derived diagnostic evidence, such as quantitative measurements and spatial observations, and their corresponding diagnostic conclusions and criteria (\textit{e.g.}, \textit{What is the cardiothoracic ratio?}), and
\textbf{\textit{2) Visual Evidence Request}}, which asks to present the image-derived diagnostic evidence directly on the image (\textit{e.g.}, highlighting anatomical regions or showing measurement overlays).
Based on the identified diagnostic task and evidence type, the agent requests the appropriate diagnostic tool call to obtain the required evidence.

\noindent
\subsection{Clinically Grounded Diagnostic Tool Execution.}
The selected diagnostic tool analyzes the chest X-ray and returns clinically grounded evidence derived from the image for the request.
For a diagnostic evidence request, the tool returns the quantitative measurements or spatial observations, and their corresponding diagnostic criteria and conclusions.
For a visual evidence request, it returns annotated chest X-rays that presents the relevant image-derived evidence
directly on the image.
The tools are implemented using CheXStruct, whose evidence-extraction, assessment, and task-specific quality-control procedures were developed with two board-certified radiologists and one board-certified radiation oncologist \citep{NEURIPS2025_7ef1f85b}.
Manual review supported reliable evidence extraction in 1,198 of 1,200 cases (99.83\%).
Because the pipeline uses rule-based geometric computations derived from these criteria, the returned evidence is deterministic and reproducible for a given image.

\noindent
\subsection{Evidence-Grounded Response Generation.}
For response generation, the LLM receives the user query, dialogue history, and tool-returned evidence, but does not directly access the chest X-ray.
It generates a response that connects the image-derived measurement or observation, the clinical criterion used to interpret it, and the resulting diagnostic conclusion.
When visual evidence is requested, the corresponding annotated chest X-ray is presented alongside the response.
This makes the diagnostic evidence trace explicit and enables users to verify the evidence supporting a conclusion throughout a multi-turn interaction.

\section{\benchmark}
\benchmark is a multi-turn dialogue benchmark designed to evaluate whether responses are correctly grounded in image-derived evidence across dialogue turns.
The benchmark is constructed as follows.

\subsection{Dialogue Scenario Definition.}
To cover diverse dialogue conditions, we define dialogue scenarios along two aspects:

\noindent
\textbf{\textit{1) Task Coverage.}}
The benchmark is restricted to the same 12 predefined diagnostic tasks as \agent.
We vary the number of diagnostic tasks addressed within a single dialogue: single-task dialogues focus on one diagnostic task, multi-task dialogues cover two diagnostic tasks, and global-to-task dialogues begin with a global question (\textit{e.g.}, overall presence of abnormalities) and then explore one to four specific diagnostic tasks in detail.

\noindent
\textbf{\textit{2) Question Flow Patterns.}} 
We define three question flow patterns to reflect different styles of diagnostic questioning. 
In the \textit{top-down} pattern, the dialogue starts with a diagnostic conclusion question and progressively requests supporting evidence (\textit{e.g.}, quantitative measurements or annotated images).
The \textit{bottom-up} pattern follows the reverse order, beginning with specific evidence and gradually leading to the diagnostic conclusion. 
In the \textit{random} pattern, questions appear in a flexible order, reflecting less structured user interactions.

\subsection{Action Sequence Construction.}
Given a dialogue scenario, we construct an action sequence to specify the turn-by-turn interaction structure of the dialogue, as in ToolDial \citep{shim2025tooldial}.
Specifically, it defines, for each turn, the type of user request, which in turn determines the expected response behavior of the model.
We consider two user request types: diagnostic evidence requests (\textit{e.g.}, measurements, spatial observations, or diagnostic conclusions) and visual evidence requests (\textit{e.g.}, highlighting anatomical regions).
For example, a sequence may follow:
\textit{Diagnostic evidence request} $\rightarrow$ \textit{Model response} $\rightarrow$ \textit{Visual evidence request} $\rightarrow$ \textit{Model response} $\rightarrow \cdots \rightarrow$ \textit{User bye} $\rightarrow$ \textit{Model bye}.
These action sequences serve as dialogue skeletons that guide the subsequent dialogue generation process.

\subsection{Dialogue Generation.}
Using the defined dialogue scenarios and action sequences, we generate multi-turn dialogues turn by turn using Gemini-3-Flash.
For each turn, we prompt the model with the dialogue scenario, the user request type assigned to that turn, and the request-specific evidence extracted from the chest X-ray by the CheXStruct pipeline \citep{NEURIPS2025_7ef1f85b}, along with the previously generated dialogue turns.
The user question is generated to be consistent with the dialogue scenario and request type and to be answerable using the provided evidence, while the response is grounded in the same evidence.
The dialogue reference evidence is derived from the same CheXStruct pipeline that \agent accesses during tool execution. This shared evidence source is intentional: it enables evaluation of whether the agent selects and faithfully communicates the evidence required by each user request throughout a multi-turn dialogue.
\benchmark is constructed from 1,200 task-specific cases spanning 504 unique chest X-rays adopted from CXReasonBench \citep{NEURIPS2025_7ef1f85b}, which were manually reviewed by a board-certified radiation oncologist to support reliable extraction of diagnostic evidence using CheXStruct.
In total, \benchmark comprises 1,946 dialogues with an average of 10.87 turns per dialogue (Table~\ref{tab:data_statistics}).
\begin{table}[htb!]
\centering
\caption{Dialogue statistics.}
\label{tab:data_statistics}
\begin{tabular}{l r}
\toprule
\textbf{Statistic} & \textbf{Value} \\
\midrule
Total dialogues & 1,946 \\
\quad Single-task & 1,200 \\
\quad Multi-task & 660 \\
\quad Global-to-task & 86 \\
\midrule
Avg. turns / dialogue & 10.87 \\
\bottomrule
\end{tabular}
\end{table}

\subsection{Dialogue Validation.}
We validate 100 randomly sampled dialogues using both an LLM-as-a-Judge \citep{zheng2023judging} (\textit{i.e.}, Gemini-3-Flash) and human evaluation conducted by 10 graduate students under the supervision of a radiation oncologist.
Dialogue quality is assessed using three criteria:
\textit{Question Compliance (0/1)}: whether each question follows the assigned diagnostic task and request type.
\textit{Answer Correctness (0/1)}: whether the response correctly addresses the question and is grounded in the provided evidence.
\textit{Naturalness (1–5)}: coherence across turns and adherence to the predefined question flow patterns.
As shown in Table~\ref{tab:data_validation}, the dialogues adhere to the intended dialogue design, provide evidence-grounded responses, and exhibit natural multi-turn interactions.

\begin{table}[htb!]
\centering
\caption{Dialogue validation results.}
\label{tab:data_validation}
\begin{tabular}{l c c}
\toprule
\textbf{Criterion} & \textbf{Gemini} & \textbf{Human} \\
\midrule
Question Compliance & 0.981 & 0.970 \\
Answer Correctness & 0.997 & 0.982 \\
Naturalness & 4.20 & 4.26 \\
\bottomrule
\end{tabular}
\end{table}

\section{Experiments}
\subsection{Evaluation Models}
We evaluate \agent with multiple LLM backbones, including the closed-source models, Gemini-3-Flash \citep{team2023gemini} and GPT-5 mini \citep{singh2025openai}, and the open-source models, Llama 3.3-70B \citep{grattafiori2024llama} and Qwen3 (4B, 8B, 32B) \citep{yang2025qwen3}.
We compare against three LVLM baselines: Gemini-3-Flash \citep{team2023gemini}, Pixtral-Large \citep{agrawal2024pixtral}, and MedGemma 27B \citep{sellergren2025medgemma}.
To comply with the MIMIC Data Use Agreement, we access Gemini through Vertex AI, deploy GPT-5 mini on Azure’s HIPAA-compliant platform, and evaluate open-source LVLMs locally. Open-source models are served with vLLM on NVIDIA A100 GPUs (four GPUs for Pixtral-Large and Llama-3.3-70B) and RTX A6000 GPUs (two GPUs for Qwen3 variants and MedGemma 27B).

We additionally evaluate a baseline termed ``Tool outputs + LLM,'' which receives the complete structured CheXStruct outputs and answers each query in a single call using the same backbone, bypassing query interpretation, tool planning, and request-specific tool execution. This comparison assesses their contribution beyond access to tool outputs.

\subsection{Evaluation Metrics}
We evaluate models using the following metrics.

\noindent
\textbf{1) Turn-level Metrics.}
Each metric is evaluated as a binary score at the turn level using an LLM-as-a-Judge \citep{zheng2023judging} (\textit{i.e.}, Gemini-3-Flash).
\textit{(i) Diagnostic Task Identification} and \textit{(ii) Evidence Type Identification} assess whether the agent correctly identifies, from the user query, the requested diagnostic task and evidence type, respectively.
These two metrics are evaluated only for {\agent}s, as they assess the tool planning stage.
\textit{(iii) Coverage} assesses whether the response fully addresses the user query, regardless of factual correctness.
\textit{(iv) Faithfulness} assesses whether a response faithfully reflects the request-specific reference evidence derived from CheXStruct, thereby evaluating tool-grounded response generation.
\textit{(v) Hallucination} is defined as the case where \textit{Coverage}=1 and \textit{Faithfulness}=0.
When evaluating LVLMs, turns requesting visual evidence are excluded, as these models do not produce visual evidence.

\noindent
\textbf{2) Dialogue-level Metrics.}
A turn is considered successful only if all applicable turn-level metrics are correct.
\textit{(i) Average Dialogue Success} is the average ratio of successful turns per dialogue.
\textit{(ii) Strict Dialogue Success} is the proportion of dialogues in which all turns are successful.

\subsection{Evaluation Settings}
We evaluate models under three complementary settings.
While \benchmark uses pre-generated user queries to ensure controlled comparison on identical questions, these fixed queries do not adapt to erroneous responses. 

\noindent
In this default setting, \textbf{\textit{1) Without GT}}, the dialogue history is constructed from the model's own outputs, allowing errors to accumulate across turns and potentially reducing interaction naturalness. 

\noindent
\textbf{\textit{2) With GT}} provides the ground-truth dialogue history at each turn, preventing error propagation and serving as an upper-bound setting under fixed user queries \citep{shim2025tooldial}.
However, because the history is corrected at every turn, this setting can mask how models behave when their own errors accumulate in interactive use. 

\noindent
\textbf{\textit{3) Dynamic User Simulator}} addresses these limitations by evaluating performance under adaptive multi-turn interactions. 
Following the same dialogue design (\textit{i.e.}, dialogue scenario and action sequence) as \benchmark, Gemini-3-Flash generates user queries conditioned on the model-generated dialogue history at each turn, allowing the queries to adapt to model responses while maintaining a similar structural pattern to the original dialogues.
\subsection{Results}

\noindent
\textbf{Importance of Clinically Grounded Diagnostic Tools.}
Tables~\ref{tab:result_static} and~\ref{tab:result_simulator} indicate that LVLMs often generate plausible responses that appear reasonable but are not correctly grounded in the image-derived diagnostic evidence.
In contrast, \agent grounds each response in diagnostic and visual evidence extracted from the chest X-ray by clinically grounded diagnostic tools, helping keep the reasoning faithfully grounded in verifiable image-derived evidence.
This advantage is further reflected in the dialogue-level metrics, where \agent achieves higher success rates, indicating more stable and coherent multi-turn diagnostic reasoning.
Overall, these results highlight the importance of integrating clinically grounded diagnostic tools for enabling reliable and verifiable evidence-grounded diagnostic reasoning.

\begin{table*}[htb!]
\centering
\caption{\textbf{Results on \benchmark under the Without GT and With GT settings.}
DTI: Diagnostic Task Identification; ETI: Evidence Type Identification; Cov: Coverage; Faith: Faithfulness; Hall: Hallucination; Avg: Average Dialogue Success; Strict: Strict Dialogue Success; Best results are underlined.
w/o: Without GT (model uses its own history); w/: With GT (ground-truth history provided)
}
\label{tab:result_static}
{\begin{adjustbox}{width=\textwidth, center}
\begin{tabular}{lcccccccccccccc}
\toprule
& \multicolumn{10}{c}{Turn-level} & \multicolumn{4}{c}{Dialogue-level} \\
\cmidrule(lr){2-11} \cmidrule(lr){12-15}
& \multicolumn{2}{c}{DTI $\uparrow$} & \multicolumn{2}{c}{ETI $\uparrow$} & \multicolumn{2}{c}{Cov $\uparrow$} & \multicolumn{2}{c}{Faith $\uparrow$} & \multicolumn{2}{c}{Hall $\downarrow$} & \multicolumn{2}{c}{Avg $\uparrow$} & \multicolumn{2}{c}{Strict $\uparrow$} \\
\cmidrule(lr){2-3} \cmidrule(lr){4-5} \cmidrule(lr){6-7} \cmidrule(lr){8-9} \cmidrule(lr){10-11} \cmidrule(lr){12-13} \cmidrule(lr){14-15}
Model & w\!/o & w\!/ & w\!/o & w\!/ & w\!/o & w\!/ & w\!/o & w\!/ & w\!/o & w\!/ & w\!/o & w\!/ & w\!/o & w\!/ \\
\midrule
\multicolumn{8}{l}{\textbf{\agent}} \\
\quad	Gemini-3-Flash	&	99.8  	&	99.8  	&	97.6  	&	97.6  	&	99.5  	&	99.4  	&	\underline{99.7}  &	\underline{99.7}  &	\underline{0.3}  &	\underline{0.3}	&	96.8 	&	96.6  &	73.4  &	72.2  	\\
\quad	GPT-5 mini	&	\underline{99.9}  	&	\underline{99.9}  	&	\underline{98.2}  	&	\underline{98.2}  	&	99.5  	&	99.5  	&	99.2  	&	99.2  	&	0.8  &	0.8  &	\underline{96.9}  	&	\underline{96.9}  	&	\underline{74.8}  	&	\underline{74.2}  	\\
\quad	Llama 3.3-70B	&	99.3  	&	99.4  	&	96.7  	&	96.9  	&	99.4  	&	99.4  	&	99.3  	&	99.2 	&	0.7  &	0.8  &	95.3  	&	95.5  	&	65.8  	&	65.2  	\\
\quad	Qwen3-32B	&	98.6  	&	97.0  	&	97.0  	&	97.3  	&	\underline{99.6}  	&	99.5  	&	97.6  	&	98.6 &	2.3  &	1.4  		&	94.2  	&	91.6  	&	61.1  	&	59.0  	\\
\quad	Qwen3-8B	&	86.3  	&	89.5  	&	86.6  	&	90.0  	&	\underline{99.6}  	&	\underline{99.7}  	&	93.7  	&	95.3  	&	6.3  &	4.6  &	76.6  	&	80.6  	&	21.1  	&	23.1  	\\
\quad	Qwen3-4B	&	91.6  	&	92.6  	&	92.1  	&	93.7  	&	\underline{99.6}  	&	99.4  	&	94.7  	&	96.3  	&	5.3  &	3.7  &	80.4  	&	82.2  	&	31.6  	&	32.9  	\\
\midrule
\multicolumn{8}{l}{\textbf{LVLM Baselines}} \\
    Gemini-3-Flash	&	-	&	-	&	-	&	-	&	98.3 	&	98.3 	&	43.1 	&	81.1 	&	55.6 	&	18.0 		&	36.9	&	69.1	&	8.6	&	9.8	\\
	Pixtral-Large	&	-	&	-	&	-	&	-	&	98.8 	&	98.8 	&	57.9 	&	79.8 	&	41.5 	&	20.1 		&	48.6	&	65.4	&	5.9	&	6.0	\\
	MedGemma 27B	&	-	&	-	&	-	&	-	&	98.9 	&	98.5 	&	53.9 	&	76.2 	&	45.4 	&	23.2 		&	46.1	&	64.1	&	6.1	&	6.5	\\
\bottomrule
\end{tabular}
\end{adjustbox}}
\end{table*}

\begin{table*}[htb!]
\centering
\caption{\textbf{Results with the Dynamic User Simulator based on the \benchmark dialogue design.} DTI: Diagnostic Task Identification; ETI: Evidence Type Identification; Cov: Coverage; Faith: Faithfulness; Hall: Hallucination; Avg: Average Dialogue Success; Strict: Strict Dialogue Success; Best results are underlined.
} 
\label{tab:result_simulator}
\begin{tabular}{lccccccc}
\toprule

& \multicolumn{5}{c}{Turn-level} & \multicolumn{2}{c}{Dialogue-level} \\
\cmidrule(lr){2-6} \cmidrule(lr){7-8}
Model &  DTI $\uparrow$ & ETI $\uparrow$ & Cov $\uparrow$ & Faith $\uparrow$ & Hall $\downarrow$ & Avg $\uparrow$ & Strict $\uparrow$ \\

\midrule
\multicolumn{5}{l}{\textbf{\agent}} \\
\quad	Gemini-3-Flash	&	98.0 	&	98.5 	&	\underline{99.9} 	&	\underline{99.9} 	&	\underline{0.1} 	&	93.3 	&	75.7 	\\
\quad	GPT-5 mini	&	\underline{99.9}	&	\underline{98.8} 	&	\underline{99.9} 	&	99.8 	&	0.2 	&	\underline{98.4} 	&	\underline{85.8} 	\\
\quad	Llama 3.3-70B	&	98.8 	&	98.3 	&	\underline{99.9} 	&	99.7 	&	0.3 	&	96.1 	&	79.9 	\\
\quad	Qwen3-32B	&	96.5 	&	97.4 	&	\underline{99.9} 	&	98.7 	&	1.3 	&	92.4 	&	67.8 	\\
\quad	Qwen3-8B	&	83.0 	&	84.7 	&	\underline{99.9} 	&	96.8 	&	3.2 	&	76.4 	&	29.4 	\\
\quad	Qwen3-4B	&	88.7 	&	89.6 	&	99.8 	&	97.3 	&	2.7 	&	81.7 	&	38.5 	\\
\midrule
\multicolumn{5}{l}{\textbf{LVLM Baselines}} \\
	Gemini-3-Flash	&	-	&	-	&	98.5 	&	46.3 	&	52.3 	&	33.8 	&	9.10 	\\
	Pixtral-Large	&	-	&	-	&	98.5 	&	48.2 	&	50.3 	&	35.5 	&	7.70 	\\
	MedGemma 27B	&	-	&	-	&	98.7 	&	44.9 	&	53.8 	&	34.8 	&	5.10 	\\

\bottomrule
\end{tabular}
\end{table*}

\noindent
\textbf{Contribution of Agent Design.}
Table~\ref{tab:result_baseline} compares \agent with the Tool outputs + LLM baseline. \agent achieves higher Faithfulness across all four evaluated backbones and higher Average and Strict Dialogue Success for three of them.
Gemini-3-Flash shows the largest gains, improving Faithfulness by 6.5 percentage points and Strict Dialogue Success by 24.7 points.
Qwen3-4B exhibits higher Coverage and Faithfulness but lower dialogue success.
These results support the additional contribution of request-specific planning and evidence selection beyond access to tool outputs.

\begin{table*}[htb!]
\centering
\caption{\textbf{Comparison with the Tool outputs + LLM baseline under the Without-GT setting.}
Since DTI and ETI are not applicable to the baseline, dialogue success uses only Coverage and Faithfulness for both methods.
Cov: Coverage; Faith: Faithfulness; Avg: Average Dialogue Success; Strict: Strict Dialogue Success.
Better results within each backbone are underlined.
}
\label{tab:result_baseline}
\begin{tabular}{llcccc}
\toprule
& & \multicolumn{2}{c}{Turn-level} & \multicolumn{2}{c}{Dialogue-level} \\
\cmidrule(lr){3-4} \cmidrule(lr){5-6}
Backbone & Method & Cov \(\uparrow\) & Faith \(\uparrow\) & Avg \(\uparrow\) & Strict \(\uparrow\) \\
\midrule
Gemini-3-Flash & \agent & \underline{99.5} & \underline{99.7} & \underline{98.9} & \underline{90.5} \\
& Tool outputs + LLM & 99.3 & 93.2 & 92.6 & 65.8 \\
\midrule
Qwen3-32B & \agent & \underline{99.6} & \underline{97.6} & \underline{95.9} & \underline{71.4} \\
& Tool outputs + LLM & \underline{99.6} & 93.2 & 92.9 & 65.4 \\
\midrule
Qwen3-8B & \agent & \underline{99.6} & \underline{93.7} & \underline{78.1} & \underline{24.7} \\
& Tool outputs + LLM & 98.5 & 93.4 & 76.6 & 21.3 \\
\midrule
Qwen3-4B & \agent & \underline{99.6} & \underline{94.7} & 82.4 & 39.7 \\
& Tool outputs + LLM & 98.6 & 90.3 & \underline{89.2} & \underline{45.6} \\
\bottomrule
\end{tabular}
\end{table*}

\noindent
\textbf{Effect of Backbone Scale and Design Flexibility.}
Tables~\ref{tab:result_static} and~\ref{tab:result_simulator} further analyze the effect of backbone scale within \agent.
While larger backbones consistently improve performance across most turn-level metrics, the gains are generally moderate, whereas more noticeable improvements are observed in dialogue-level success.
This suggests that increased model capacity primarily enhances query interpretation accuracy and multi-turn reasoning stability.
Importantly, even small backbones (\textit{e.g.}, Qwen3-4B and 8B) already surpass all LVLMs across all metrics, indicating that the primary performance gains arise from the agent design, namely clinically grounded diagnostic tool-based evidence grounding, rather than from model scale alone.
These results highlight the cost-efficiency and flexibility of the proposed agent design: the core diagnostic reasoning capability is largely preserved across backbone sizes, allowing practitioners to flexibly trade off performance and computational cost by swapping the underlying language model without redesigning the system.

\noindent
\textbf{Robust Evidence Grounding Across Evaluation Settings.}
In Table~\ref{tab:result_static}, \agent maintains strong performance in both Without GT and With GT settings, indicating robust evidence-grounded reasoning even when relying on its own generated dialogue history, where errors can accumulate across turns.
In contrast, LVLMs show substantial gains in faithfulness and reduced hallucination when ground-truth history is provided, suggesting that they may leverage diagnostic evidence already present in the ground-truth history rather than consistently grounding their responses in image-derived evidence.
Under the dynamic user simulator (Table~\ref{tab:result_simulator}), \agent continues to outperform LVLMs by a similar margin, 
demonstrating that its evidence grounding remains stable under response-adaptive interactions.
Overall, these results show that \agent maintains reliable evidence-grounded reasoning across both controlled fixed-query evaluations and response-adaptive multi-turn interactions.

\noindent
\textbf{Evaluation with an Independent Judge.}
We evaluated 100 dialogues per backbone for Gemini-3-Flash, Llama-3.3-70B, and Qwen3-8B using GPT-6 Astra. 
The GPT-6 Astra evaluation supports the same ranking of the three backbones, but the lower scores for Qwen3-8B show that absolute performance estimates remain sensitive to judge choice.
Detailed results are provided in Table~\ref{tab:result_judge}.

\noindent
\textbf{Confidence Intervals.}
We computed 95\% confidence intervals under the Dynamic User Simulator setting using 10,000 image-level cluster-bootstrap resamples. The intervals remained separated between \agent and the LVLM baselines across all shared metrics, supporting the stability of the performance advantage under resampling. Detailed procedures and results are provided in Table~\ref{tab:result_CI}.

\section{Conclusion}
We presented \agent, a tool-augmented diagnostic agent that combines an LLM with clinically grounded diagnostic tools to provide evidence-grounded diagnostic reasoning through multi-turn dialogue.
We also introduced
\benchmark, a multi-turn chest X-ray benchmark for evaluating whether an agent selects and faithfully communicates the diagnostic and visual evidence required by user requests across 12 diagnostic tasks.
Across multiple LLM backbones, the results show that \agent can generate responses consistent with the corresponding tool-derived evidence throughout a dialogue, while the same-evidence comparison supports the additional contribution of request-specific planning and evidence selection.
By linking diagnostic conclusions to image-derived measurements or spatial observations, clinical criteria, and requested visual annotations, \agent enables users to verify how a diagnostic conclusion is derived.
The current scope is limited to 12 diagnostic tasks on chest X-rays.
The agent may propagate incorrect tool evidence, and its ability to detect such errors has not been established.
Future work will investigate evidence verification and appropriate abstention, and extend the framework to broader chest X-ray assessments and other healthcare modalities, such as ECG, with domain-specific validation of the tools and resulting interactions.

\acks{
This work was supported by the Institute of Information \& Communications Technology Planning \& Evaluation (IITP) grants (No.RS-2019-II190075), the Korea Health Industry Development Institute (KHIDI) grant (No.RS-2025-02213750), and the National Research Foundation of Korea (NRF) grant (NRF-2020H1D3A2A03100945, RS-2026-25484088), and the Medical Scientist Training Program funded by the Korean government (MSIT, MOHW).
}

\clearpage
\bibliography{ref}

\clearpage
\appendix
\section{Diagnostic Tasks}\label{apd:diagnostic_tasks}

Tables~\ref{tab:diagnostic_tasks_thoracic}, ~\ref{tab:diagnostic_tasks_airway}, and ~\ref{tab:diagnostic_tasks_quality} summarize the definitions and assessment criteria of the 12 diagnostic tasks supported by \agent.

\section{Agent Prompts}\label{apd:agent_prompts}

This section presents the prompts used in the three LLM-based stages of \agent: query interpretation, tool planning, and evidence-grounded response generation.
The clinically grounded diagnostic tools are deterministic and therefore do not use an LLM prompt.

\subsection{Query Interpretation Prompt}
\label{app:query_interpretation_prompt}

\noindent
Given the current user query and the prior conversation history, the following prompt maps the user query to one or more supported diagnostic tasks or a special category.

\begin{quote}
\small\ttfamily
\noindent
[Role]\par
You are a classifier that maps a user's chest X-ray-related question to one
or more [Available Diagnostic Tasks], or to a [Special Category], based on
the user's intent and the prior conversation context. Your classification MUST
reflect whether the user's question can be answered using only chest
X-ray-derived diagnostic information, as defined in the [Definition:
Chest X-ray-Derived Diagnostic Information] section below.\par

[Definition: Chest X-ray-Derived Diagnostic Information]\par
For the purpose of task classification, a question is considered in scope ONLY
if it can be answered using diagnostic information that can be directly derived from a chest X-ray image itself.\par

Chest X-ray-derived diagnostic information includes:\par
- Diagnostic conclusions (e.g., normal, abnormal, enlarged, adequate, PA, or AP)\par
- Quantitative or categorical results (e.g., measurements, ratios, labels)\par
- Definitions or meanings of diagnostic findings\par
- How diagnostic findings are assessed or evaluated (e.g., formulas,
thresholds, decision rules)\par
- Case-specific diagnostic questions (e.g., values or labels in this specific image)\par
- General diagnostic questions (e.g., what a finding means, how it is assessed)\par
- Visualization of anatomical regions or measurement procedures used to
assess a diagnostic finding\par

Any request beyond the above scope MUST be treated as Out of Scope, even if a
supported diagnostic task is mentioned.\par

[Rules]\par
1. Conversation continuity\par
(1) If there is one or more active diagnostic task(s) inferred from the prior
conversation context, interpret the current question in the context of those
active task(s) FIRST.\par
(2) If there is an active diagnostic task inferred from the prior conversation
context, and the current question does NOT explicitly introduce a different
diagnostic finding or condition, the question MUST be classified as the active
diagnostic task.\par
(3) Under the condition of Rule (2), requests to show, locate, or visualize
anatomical regions, or to ask for a quantitative assessment, measurement, or
diagnostic label on the chest X-ray MUST NOT be treated as introducing a new
diagnostic task, even if those anatomical regions are also used in other
diagnostic tasks.\par

2. Explicit diagnostic task mention\par
If the user clearly and explicitly asks about a diagnostic task listed in
[Available Diagnostic Tasks], classify the question as that diagnostic task,
ONLY IF the requested information falls within the scope of chest
X-ray-derived diagnostic information defined above.\par

3. Multiple diagnostic tasks\par
Assign multiple diagnostic tasks ONLY if the user explicitly asks about
multiple DIFFERENT diagnostic findings or conditions in the same question.
Requests to visualize, locate, or measure multiple anatomical structures that
are used together to assess a SINGLE diagnostic task MUST be classified as
that diagnostic task ONLY and MUST NOT introduce additional diagnostic tasks.\par

[Available Diagnostic Tasks]\par
[AVAILABLE DIAGNOSTIC TASKS]\par

[Special Categories]\par
- Overall Findings: Use this category ONLY when the user explicitly asks about
the overall or global radiological findings of the chest X-ray as a whole.
The question must focus on a comprehensive radiological assessment rather than
on any single specific diagnostic finding.\par
- Overall Quality: Use this category ONLY when the user explicitly asks about
the overall image quality of the chest X-ray. The question must focus on how
the image was acquired or how suitable it is for interpretation.\par
- Out of Scope: Use Out of Scope ONLY when the requested diagnostic task is
not included in [Available Diagnostic Tasks], or when the requested information
cannot be derived from a chest X-ray image alone, including patient history,
laboratory results, treatment plans, prognosis, or additional imaging studies.\par
- Exit: Use this category when the user explicitly indicates that they have no
more questions or are done.\par

[Output Format (STRICT)]\par
Return a JSON object that conforms to the required response schema. Populate
``tasks'' with the selected diagnostic task(s) or special category.
\end{quote}

\noindent
At runtime, the \texttt{[Available Diagnostic Tasks]} section is populated
with the definitions and assessment methods of all supported diagnostic tasks.
The following is an excerpt of the actual inserted content.

\begin{quote}
\small\ttfamily
- Cardiomegaly:\par
\quad Definition: Cardiomegaly refers to an enlargement of the cardiac
silhouette on a chest X-ray.\par
\quad Assessment Method: Cardiomegaly is assessed using the cardiothoracic
ratio (CTR), which compares the width of the heart to the width of the lungs.
When CTR is below 0.50, the diagnosis is ``normal''. When CTR is 0.55 or
greater, the diagnosis is ``enlarged''. When CTR is at or above 0.50 but
below 0.55, the diagnosis is determined by the chest X-ray
projection-specific threshold: 0.50 for PA and 0.55 for AP. A CTR meeting or
exceeding the projection-specific threshold is ``enlarged''; otherwise,
``normal''.
\end{quote}

\noindent
The same format is used for all supported diagnostic tasks.
\subsection{Tool Planning Prompt}
\label{app:tool_planning_prompt}

\noindent
Given the user query and the diagnostic task(s) identified during query
interpretation, the following prompt selects the required information type for
each task.

\begin{quote}
\small\ttfamily
\noindent
[Role]\par
You are an information-type selector for chest X-ray diagnostic tasks.
Your job is to choose exactly ONE information type for EACH listed task so the
system can retrieve the right data. Do NOT answer the user's question. Do NOT
request additional info.\par

[Input]\par
- You will see the user's question.\par
- You will see one or more diagnostic tasks (task\_name), each with its
definition.\par
- For EACH task, select the information type needed to answer ONLY the parts
of the question that refer to that task.\par

[Information Types]\par
1. diagnostic\_evidence\par
Use this type when the user asks about diagnostic information related to a
supported finding, including:\par
- Diagnostic conclusions (e.g., normal, abnormal, enlarged)\par
- Quantitative or categorical results (e.g., measurement values, ratios, status)\par
- Definitions or meanings of a diagnostic task\par
- How a diagnostic finding is assessed or evaluated (e.g., formulas,
thresholds, decision rules)\par

2. visual\_evidence\par
Use ONLY when the user explicitly asks to show, locate, highlight, or visually
demonstrate anatomy, landmarks, or measurement markings on the chest X-ray
(e.g., ``show me where the heart is'', ``show how you measured it'').\par

[Selection Rules]\par
- Choose the information type needed to answer only the parts of the question
that refer to each task.
\end{quote}
\subsection{Evidence-Grounded Response Generation Prompt}
\label{app:response_generation_prompt}

\noindent
For each selected diagnostic task, the tool populates
\texttt{[PROVIDED INFORMATION]} with the task definition and the requested
diagnostic or visual evidence. The visualization rules below are included only
when visual evidence is returned by the tool.

\begin{quote}
\small\ttfamily
\noindent
[Role]\par
You are a chest X-ray diagnostic agent. Your role is to answer the user's
question by synthesizing information from [Provided Information] below, using
only what is necessary to address the question. You must behave as if you
personally reviewed the chest X-ray image, performed the measurements, and
made the assessment yourself.\par

[Provided Information]\par
[PROVIDED INFORMATION]\par

[Core Answering Rules]\par
- Base the answer on the information contained in [Provided Information], using
only what is relevant and necessary to address the user's question.\par
- You MUST speak as if you are directly reviewing the chest X-ray image
yourself, performing the measurements, and making the assessment based on your
own observation of the image (e.g., do NOT say ``based on the provided data''
or ``according to the measurements above'').\par
- Do NOT mention or imply that the information was ``provided'', ``retrieved'',
``available'', ``given'', or obtained from any external source.\par

[Visualization Rules; included only when visual evidence is available]\par
- The [Provided Information] includes visualization content for a diagnostic
task. Therefore the answer MUST explicitly include the corresponding image
path(s) using the format: [Figure: \textless image path\textgreater].\par
- When referencing a visualization, the answer MUST describe the image as if
you are directly viewing it, and MUST guide the user on what to look for in
the displayed image, such as highlighted anatomy, landmarks, or measurement
lines.\par
- The explanation of the visualization MUST rely only on the descriptions and
contents explicitly provided in the [Provided Information], without adding new
annotations, alternative markings, or hypothetical visual features.\par

[Diagnostic Conclusion Rules]\par
- If the user's question asks for a diagnostic conclusion (e.g., presence,
absence, or overall assessment of a finding), you MUST clearly state the
conclusion using the relevant labels or measurements in [Provided Information].\par

[Answer Style]\par
- State findings as direct observations from the image.\par
- When limitations apply, state them clearly and concisely without speculation.\par
- Do NOT hedge unnecessarily when image quality is adequate.\par

[Output Requirement]\par
Produce a single coherent answer that addresses the user's question using only
the allowed information and rules above.
\end{quote}

\noindent
At runtime, the \texttt{[Provided Information]} section is populated with
the definition of each selected diagnostic task and the corresponding
tool-returned information. For a \texttt{diagnostic\_evidence} query, the
tool returns the task explanation and measurement values. For a
\texttt{visual\_evidence} query, it returns visualization descriptions and
their corresponding image paths. The following examples show the inserted
content format; numeric values and image paths are illustrative.\\

\noindent\textit{Diagnostic-evidence query.}
\begin{quote}
\small\ttfamily
\noindent
- Cardiomegaly: Cardiomegaly refers to an enlargement of the cardiac
silhouette on a chest X-ray.\par
\quad \{`diagnostic\_evidence': \{\par
\quad\quad `Explanation': \{\par
\quad\quad\quad `assessment\_method': \{\par
\quad\quad\quad\quad `description': `Cardiomegaly is assessed using the cardiothoracic ratio (CTR), which compares the width of the heart to the width of the lungs. When CTR is below 0.50, the diagnosis is normal. When CTR is 0.55 or greater, the diagnosis is enlarged. When CTR is at or above 0.50 but below 0.55, the diagnosis is determined by the chest X-ray projection-specific threshold: 0.50 for PA and 0.55 for AP. A CTR meeting or exceeding the projection-specific threshold is enlarged; otherwise normal.',\par
\quad\quad\quad\quad `assessment\_basis': `CTR = heart\_width /
lung\_width',\par
\quad\quad\quad\quad `threshold': \{\par
\quad\quad\quad\quad\quad `PA (posteroanterior)': 0.50,\par
\quad\quad\quad\quad\quad `AP (anteroposterior)': 0.55\par
\quad\quad\quad\quad \}\par
\quad\quad\quad \}\par
\quad\quad \},\par
\quad\quad `Measurement': \{\par
\quad\quad\quad `heart\_width': \{\par
\quad\quad\quad\quad `description': `The maximal horizontal width of the
cardiac silhouette measured on the chest X-ray (in pixels).',\par
\quad\quad\quad\quad `value': 250\par
\quad\quad\quad \},\par
\quad\quad\quad `lung\_width': \{\par
\quad\quad\quad\quad `description': `The maximal horizontal width of the lungs measured below the superior border of the heart (in pixels).',\par
\quad\quad\quad\quad `value': 500\par
\quad\quad\quad \},\par
\quad\quad\quad `CTR': \{\par
\quad\quad\quad\quad `description': 'The cardiothoracic ratio, defined as
the ratio of heart width to lung width.',\par
\quad\quad\quad\quad `value': 0.50\par
\quad\quad\quad \}\par
\quad\quad \}\par
\quad \}\}
\end{quote}

\noindent\textit{Visual-evidence query.}
\begin{quote}
\small\ttfamily
\noindent
- Cardiomegaly: Cardiomegaly refers to an enlargement of the cardiac
silhouette on a chest X-ray.\par
\quad \{`visual\_evidence': \{\par
\quad\quad `heart': \{\par
\quad\quad\quad `description': `A chest X-ray image in which the cardiac
silhouette is highlighted in red to indicate the anatomical extent of the
heart.',\par
\quad\quad\quad `value': `example/cardiomegaly\_heart.png'\par
\quad\quad \},\par
\quad\quad `lungs': \{\par
\quad\quad\quad `description': `A chest X-ray image in which the lungs are
highlighted in blue to illustrate the lung regions used for measurement.',\par
\quad\quad\quad `value': `example/cardiomegaly\_lungs.png'\par
\quad\quad \},\par
\quad\quad `measurement': \{\par
\quad\quad\quad `description': `A chest X-ray image displaying measurement
overlays, where the heart width is marked with a red line and the lung width
is marked with a blue line, corresponding to the cardiothoracic ratio
calculation.',\par
\quad\quad\quad `value': `example/cardiomegaly\_measurement.png'\par
\quad\quad \}\par
\quad \}\}
\end{quote}

\begin{table*}[htb!]
\centering
\scriptsize
\caption{Thoracic diagnostic tasks supported by \agent.}
\label{tab:diagnostic_tasks_thoracic}
\begin{tabular}{@{}>{\raggedright\arraybackslash}p{0.18\textwidth}>{\raggedright\arraybackslash}p{0.30\textwidth}>{\raggedright\arraybackslash}p{\dimexpr0.52\textwidth-4\tabcolsep\relax}@{}}
\toprule
\textbf{Task} & \textbf{Definition} & \textbf{Assessment criteria} \\
\midrule
Cardiomegaly &
Enlargement of the cardiac silhouette on a chest X-ray. &
Assessed using the cardiothoracic ratio (CTR), which compares heart width with lung width. A CTR below 0.50 is normal and a CTR of 0.55 or greater is enlarged. For CTR values from 0.50 to 0.55, the projection-specific threshold is 0.50 for PA and 0.55 for AP images. \\

Mediastinal widening &
Abnormally increased width of the mediastinum on a chest X-ray. &
Assessed using the ratio of maximal upper-mediastinal width to lung width measured at the same horizontal level. For borderline values, the projection-specific threshold is 0.25 for PA and 0.33 for AP images. \\

Aortic knob enlargement &
Increased size of the aortic knob on a chest X-ray. &
Assessed by measuring the maximal aortic-knob width and comparing it with the median tracheal width to form a ratio. \\

Ascending aorta enlargement &
Abnormal widening of the ascending thoracic aorta on a chest X-ray. &
Assessed visually by determining whether the ascending aorta extends beyond a reference line drawn between the right heart border and the inner margin of the right lung. \\

Descending aorta enlargement &
Abnormal widening of the descending thoracic aorta on a chest X-ray. &
Assessed by measuring the maximal descending-aorta width and comparing it with the median tracheal width to form a ratio. \\

Descending aorta tortuosity &
Abnormal curvature or twisting of the descending aorta on a chest X-ray. &
Assessed by computing curvature at six annotated points along the descending aorta, which is divided into five equal-length sections. Curvature at these points is calculated using finite-difference methods. \\
\bottomrule
\end{tabular}
\end{table*}

\begin{table*}[htb!]
\centering
\scriptsize
\caption{Airway diagnostic tasks supported by \agent.}
\label{tab:diagnostic_tasks_airway}
\begin{tabular}{@{}>{\raggedright\arraybackslash}p{0.18\textwidth}>{\raggedright\arraybackslash}p{0.30\textwidth}>{\raggedright\arraybackslash}p{\dimexpr0.52\textwidth-4\tabcolsep\relax}@{}}
\toprule
\textbf{Task} & \textbf{Definition} & \textbf{Assessment method} \\
\midrule
Tracheal deviation &
Displacement of the trachea from the thoracic midline on a chest X-ray. &
Assessed by visually comparing the position of the trachea with the thoracic midline, which is approximated using the vertebral spinous processes. \\

Carina angle &
The angle at the tracheal bifurcation where the trachea divides into the right and left main bronchi on a chest X-ray. &
Assessed by identifying the bifurcation point and the axes of the right and left main bronchi, then measuring the angle formed at the bifurcation. \\
\bottomrule
\end{tabular}
\end{table*}

\begin{table*}[htb!]
\centering
\scriptsize
\caption{Chest X-ray image-quality assessment tasks supported by \agent.}
\label{tab:diagnostic_tasks_quality}
\begin{tabular}{@{}>{\raggedright\arraybackslash}p{0.18\textwidth}>{\raggedright\arraybackslash}p{0.30\textwidth}>{\raggedright\arraybackslash}p{\dimexpr0.52\textwidth-4\tabcolsep\relax}@{}}
\toprule
\textbf{Task} & \textbf{Definition} & \textbf{Assessment method} \\
\midrule
Inspiration &
Degree of lung expansion at the time of chest X-ray acquisition, which affects image quality and interpretability. &
Assessed by counting the right posterior ribs along the mid-clavicular line and identifying the rib immediately above the right hemidiaphragm. \\

Inclusion &
Whether the chest X-ray captures all relevant thoracic anatomy needed for reliable interpretation. &
Assessed by visually confirming that the lung apices, lateral rib margins, and both costophrenic angles are fully visible. \\

Rotation &
Patient alignment during chest X-ray acquisition, assessed through clavicular symmetry relative to the spinous-process midline. &
Assessed by measuring the horizontal distances from the medial ends of the left and right clavicles to the interpolated midline derived from spinous-process points at the same vertical level. \\

Projection &
Orientation of the X-ray beam relative to the patient, commonly categorized as anteroposterior (AP) or posteroanterior (PA). &
Assessed by measuring the overlap between each scapula and its corresponding lung field. The overlap ratio compares the overlapping area with the total scapular area. \\
\bottomrule
\end{tabular}
\end{table*}

\section{Additional Experiments}
\subsection{Evaluation with an Independent Judge}
\label{sec:judge_evaluation}

We evaluated Coverage and Faithfulness using GPT-6 Astra on 100 dialogues from each of Gemini-3-Flash, Llama-3.3-70B, and Qwen3-8B, totaling 300 dialogues. We used the same evaluation prompts and recomputed dialogue-success metrics. For direct comparison, Gemini-3-Flash judge scores were computed on the same dialogues.

Table~\ref{tab:result_judge} shows similar scores for Gemini-3-Flash and Llama-3.3-70B across judges, but lower Faithfulness and dialogue success for Qwen3-8B under GPT-6 Astra. Both judges therefore support the same dialogue-success ranking among the three backbones, while the differences for Qwen3-8B indicate sensitivity to judge choice.

\begin{table*}[htb!]
\centering
\caption{\textbf{Comparison of Gemini-3-Flash and GPT-6 Astra judges on matched dialogues.}
Each backbone is evaluated on 100 dialogues using identical evaluation prompts.
Cov: Coverage; Faith: Faithfulness; Avg: Average Dialogue Success; Strict: Strict Dialogue Success.
}
\label{tab:result_judge}
\begin{tabular}{llcccc}
\toprule
& & \multicolumn{2}{c}{Turn-level} & \multicolumn{2}{c}{Dialogue-level} \\
\cmidrule(lr){3-4} \cmidrule(lr){5-6}
Backbone & Judge & Cov \(\uparrow\) & Faith \(\uparrow\) & Avg \(\uparrow\) & Strict \(\uparrow\) \\
\midrule
Gemini-3-Flash & Gemini-3-Flash & 99.9 & 99.8 & 98.1 & 75.0 \\
& GPT-6 Astra & 100.0 & 99.7 & 98.1 & 75.0 \\
\midrule
Llama-3.3-70B & Gemini-3-Flash & 99.9 & 99.7 & 97.8 & 76.0 \\
& GPT-6 Astra & 100.0 & 99.3 & 97.5 & 74.0 \\
\midrule
Qwen3-8B & Gemini-3-Flash & 99.9 & 97.5 & 86.9 & 42.0 \\
& GPT-6 Astra & 99.8 & 91.7 & 81.9 & 31.0 \\
\bottomrule
\end{tabular}
\end{table*}

\subsection{Confidence Intervals}
\label{sec:confidence_intervals}

We computed 95\% confidence intervals for all metrics under the Dynamic User Simulator setting using 10,000 image-level cluster-bootstrap resamples with a random seed of 42.
Each resample draws the original number of unique chest X-rays with replacement, retaining all associated dialogues and turns together.
Images sampled multiple times contribute repeated copies of their associated results. We recompute each metric and use the 2.5th and 97.5th percentiles as the interval bounds.
Table~\ref{tab:result_CI} reports the estimates and confidence intervals. The intervals remain separated between \agent and the LVLM baselines across all shared metrics, supporting the stability of the observed performance advantage under image-level resampling within this benchmark.

\begin{table*}[htb!]
\centering
\caption{\textbf{95\% confidence intervals under the Dynamic User Simulator setting.}
Values are percentages, shown as estimate [95\% CI], computed using 10,000 image-level cluster-bootstrap resamples.
DTI: Diagnostic Task Identification; ETI: Evidence Type Identification; Cov: Coverage; Faith: Faithfulness; Hall: Hallucination; Avg: Average Dialogue Success; Strict: Strict Dialogue Success.
}
\label{tab:result_CI}
{\begin{adjustbox}{width=\textwidth, center}
\begin{tabular}{lccccccc}
\toprule
& \multicolumn{5}{c}{Turn-level} & \multicolumn{2}{c}{Dialogue-level} \\
\cmidrule(lr){2-6} \cmidrule(lr){7-8}
Model & DTI \(\uparrow\) & ETI \(\uparrow\) & Cov \(\uparrow\) & Faith \(\uparrow\) & Hall \(\downarrow\) & Avg \(\uparrow\) & Strict \(\uparrow\) \\
\midrule
\multicolumn{8}{l}{\textbf{\agent}} \\
\quad Gemini-3-Flash
& 98.0 [97.6, 98.3]
& 98.5 [98.3, 98.7]
& 99.9 [99.8, 99.9]
& 99.9 [99.9, 100.0]
& 0.1 [0.1, 0.2]
& 93.4 [92.6, 94.1]
& 75.7 [73.7, 77.7] \\
\quad GPT-5 mini
& 99.9 [99.8, 99.9]
& 98.8 [98.6, 99.0]
& 99.9 [99.9, 100.0]
& 99.8 [99.7, 99.8]
& 0.3 [0.2, 0.3]
& 98.4 [98.2, 98.6]
& 85.8 [84.4, 87.3] \\
\quad Llama 3.3-70B
& 98.8 [98.5, 99.0]
& 98.3 [98.1, 98.5]
& 99.9 [99.8, 99.9]
& 99.7 [99.6, 99.8]
& 0.3 [0.2, 0.4]
& 96.2 [95.6, 96.7]
& 80.0 [78.2, 81.7] \\
\quad Qwen3-32B
& 96.5 [96.1, 96.9]
& 97.4 [97.1, 97.7]
& 99.9 [99.9, 100.0]
& 98.7 [98.5, 98.9]
& 1.3 [1.1, 1.5]
& 92.4 [91.6, 93.1]
& 67.8 [65.9, 69.8] \\
\quad Qwen3-8B
& 83.0 [82.2, 83.7]
& 84.7 [83.9, 85.5]
& 99.9 [99.8, 99.9]
& 96.8 [96.5, 97.2]
& 3.2 [2.8, 3.5]
& 76.4 [75.5, 77.4]
& 29.4 [27.5, 31.5] \\
\quad Qwen3-4B
& 88.7 [88.0, 89.4]
& 89.6 [88.9, 90.3]
& 99.8 [99.8, 99.9]
& 97.3 [97.0, 97.6]
& 2.7 [2.4, 3.0]
& 81.7 [80.7, 82.6]
& 38.5 [36.2, 40.8] \\
\midrule
\multicolumn{8}{l}{\textbf{LVLM Baselines}} \\
Gemini-3-Flash
& -- & --
& 98.5 [98.1, 98.9]
& 46.3 [43.7, 48.7]
& 52.3 [49.7, 55.0]
& 33.8 [31.8, 35.8]
& 9.1 [7.8, 10.4] \\
Pixtral-Large
& -- & --
& 98.5 [98.1, 98.8]
& 48.2 [46.4, 49.9]
& 50.3 [48.6, 52.1]
& 35.5 [34.1, 36.9]
& 7.7 [6.6, 8.9] \\
MedGemma 27B
& -- & --
& 98.7 [98.4, 98.9]
& 44.9 [43.2, 46.5]
& 53.8 [52.2, 55.4]
& 34.8 [33.4, 36.2]
& 5.1 [4.2, 6.2] \\
\bottomrule
\end{tabular}
\end{adjustbox}}
\end{table*}

\subsection{Per-Task Results}
\label{sec:per_task_results}

Table~\ref{tab:result_per_task} reports per-task performance under the Dynamic User Simulator setting for Gemini-3-Flash and Qwen3-32B. Each task appears in 221--241 dialogues. Metrics are computed using only turns associated with the corresponding task. Average Dialogue Success is the mean proportion of successful task-specific turns per dialogue, while Strict Dialogue Success requires all task-specific turns within a dialogue to succeed.
Gemini-3-Flash maintains high Coverage and Faithfulness across all 12 tasks. Qwen3-32B exhibits greater variation, with Faithfulness ranging from 95.6\% for cardiomegaly to 99.7\% for carina angle and descending aorta tortuosity. Its Strict Dialogue Success ranges from 57.9\% for cardiomegaly to 87.7\% for carina angle, showing that high aggregate response-level scores can coexist with substantial task-specific differences in dialogue success.

\subsection{Tool-Only Template Baseline}
\label{app:template_baseline}

We evaluated a CheXStruct + fixed-template baseline on 100 randomly sampled dialogues from \benchmark. Annotated diagnostic tasks and evidence types select the corresponding tool outputs, which are rendered through fixed templates without LLM-based question interpretation or response generation. This is therefore an oracle-selection baseline rather than an end-to-end conversational system. Since DTI and ETI are not applicable, dialogue success is computed using Coverage and Faithfulness only.

The baseline achieved 97.4\% Coverage, 100.0\% Faithfulness, 97.4\% Average Dialogue Success, and 74.0\% Strict Dialogue Success. Coverage failures included three types of incomplete responses:

\begin{itemize}
\item \textbf{Overall assessments.} When asked whether the image was interpretable overall, the template listed inclusion, rotation, inspiration, and projection results without providing an overall conclusion.
\item \textbf{Additional explanations.} When asked how inadequate inspiration affects interpretation, the template provided the rib count and ``inadequate'' label without explaining the implications.
\item \textbf{Combined requests.} When asked whether the aorta was enlarged and to show supporting visual evidence, the template returned visual evidence but omitted the diagnostic conclusion.
\end{itemize}

These results show that structured tool outputs support many questions when the correct task and evidence type are supplied. However, reproducing these outputs does not always provide a complete, question-specific answer. \agent additionally interprets user queries and dialogue context, selects relevant evidence, and generates tailored explanations.

\subsection{Unsupported-Question Evaluation}
\label{sec:unsupported_questions}

We evaluated \agent with Gemini-3-Flash on 100 single-turn questions unsupported by the available tools. These comprised 50 out-of-scope questions concerning unsupported findings, treatment, management, or prognosis, and 50 questions requiring unavailable information, such as prior examinations, clinical history, or additional test results. Examples include Does this chest X-ray show pneumonia?'', Should this patient start a diuretic based on this chest X-ray?'', and ``Has the cardiac silhouette increased in size since the previous examination?''

The agent declined all 50 out-of-scope requests and withheld the requested conclusions for all 50 questions requiring unavailable information. No response subsequently provided the requested unsupported patient-specific conclusion. However, responses predominantly invoked scope restrictions. These results demonstrate refusal of unsupported requests, rather than reliable detection of insufficient evidence within supported tasks. Systematic evaluation of ambiguous cases and incomplete tool evidence remains future work.

\subsection{Qualitative Analysis of Evidence-Selection Errors}
\label{sec:error_analysis}

We examined turns with incorrect diagnostic-task or evidence-type identification by reviewing the user query, dialogue history, retrieved evidence, and generated response. This analysis concerns errors in selecting evidence, rather than inaccuracies in the underlying tool measurements.

In one case, Gemini-3-Flash selected visual evidence for a quantitative measurement request, then generated numerical values and units unsupported by either the retrieved evidence or the preceding dialogue. In another, Qwen3-8B explicitly recognized that the retrieved evidence concerned cardiomegaly rather than the requested mediastinal assessment, but continued with unsupported thresholds and interpretations instead of withholding an answer.

These cases show that evidence-selection errors can lead to unsupported responses, even when the agent recognizes the mismatch. The current framework lacks a dedicated mechanism to verify evidence sufficiency and ensure appropriate abstention. Future work should investigate such safeguards and evaluate robustness to deliberately incorrect, missing, or contradictory tool outputs.

\begin{table*}[htb!]
\centering
\caption{\textbf{Per-task results under the Dynamic User Simulator setting.}
Metrics are computed over task-specific turns within each dialogue.
Cov: Coverage; Faith: Faithfulness; Hall: Hallucination; Avg: Average Dialogue Success; Strict: Strict Dialogue Success.
}
\label{tab:result_per_task}
{\begin{adjustbox}{width=0.7\textwidth, center}
\begin{tabular}{lccccc}
\toprule
& \multicolumn{3}{c}{Turn-level} & \multicolumn{2}{c}{Dialogue-level} \\
\cmidrule(lr){2-4} \cmidrule(lr){5-6}
Task & Cov \(\uparrow\) & Faith \(\uparrow\) & Hall \(\downarrow\) & Avg \(\uparrow\) & Strict \(\uparrow\) \\
\midrule
\multicolumn{6}{l}{\textbf{Gemini-3-Flash}} \\
\quad Aortic knob enlargement & 99.7 & 99.9 & 0.1 & 95.4 & 79.9 \\
\quad Ascending aorta enlargement & 99.6 & 100.0 & 0.0 & 87.7 & 75.4 \\
\quad Cardiomegaly & 100.0 & 99.9 & 0.1 & 91.8 & 81.8 \\
\quad Carina angle & 99.9 & 99.9 & 0.1 & 91.1 & 77.1 \\
\quad Descending aorta enlargement & 99.9 & 100.0 & 0.0 & 88.2 & 71.1 \\
\quad Descending aorta tortuosity & 99.9 & 99.9 & 0.1 & 95.8 & 86.3 \\
\quad Inclusion & 100.0 & 100.0 & 0.0 & 92.1 & 82.8 \\
\quad Inspiration & 100.0 & 99.7 & 0.1 & 92.1 & 78.6 \\
\quad Mediastinal widening & 99.9 & 100.0 & 0.0 & 95.3 & 89.8 \\
\quad Projection & 99.9 & 99.4 & 0.5 & 94.3 & 75.3 \\
\quad Rotation & 99.9 & 99.9 & 0.1 & 93.1 & 82.0 \\
\quad Tracheal deviation & 99.8 & 100.0 & 0.0 & 91.8 & 75.1 \\
\midrule
\multicolumn{6}{l}{\textbf{Qwen3-32B}} \\
\quad Aortic knob enlargement & 99.7 & 98.3 & 1.7 & 91.6 & 68.5 \\
\quad Ascending aorta enlargement & 99.9 & 99.0 & 1.0 & 95.5 & 82.0 \\
\quad Cardiomegaly & 100.0 & 95.6 & 4.4 & 90.9 & 57.9 \\
\quad Carina angle & 100.0 & 99.7 & 0.4 & 96.8 & 87.7 \\
\quad Descending aorta enlargement & 99.8 & 98.7 & 1.3 & 94.4 & 70.7 \\
\quad Descending aorta tortuosity & 99.9 & 99.7 & 0.3 & 94.1 & 74.3 \\
\quad Inclusion & 99.9 & 98.9 & 1.1 & 91.2 & 80.1 \\
\quad Inspiration & 99.9 & 98.4 & 1.6 & 91.8 & 75.9 \\
\quad Mediastinal widening & 100.0 & 96.9 & 3.1 & 92.1 & 72.5 \\
\quad Projection & 100.0 & 99.3 & 0.7 & 90.3 & 64.2 \\
\quad Rotation & 99.9 & 99.6 & 0.4 & 95.6 & 81.9 \\
\quad Tracheal deviation & 99.9 & 99.5 & 0.5 & 87.8 & 77.9 \\
\bottomrule
\end{tabular}
\end{adjustbox}}
\end{table*}

\subsection{Computational Efficiency}
\label{app:efficiency}

We report approximate average wall-clock processing times per dialogue in Table~\ref{tab:latency}. Times were estimated from intervals between successive dialogue-result files and include diagnostic tool execution and all assistant responses. Tool outputs are reused throughout each dialogue, avoiding repeated image analysis. These measurements describe total dialogue processing time rather than per-response latency.
For Gemini-3-Flash, average token usage per single-task dialogue was 26,185 prompt tokens and 1,072 response tokens for \agent, compared with 23,881 prompt tokens and 1,309 response tokens for the direct LVLM baseline. Agent counts include all three LLM calls per turn, whereas the LVLM baseline uses one call and receives the original CXR as input.
Estimated inference API costs were approximately \$47 per backbone for a full run of 1,946 dialogues in each evaluation setting for Gemini-3-Flash and GPT-5 mini. These are approximate costs rather than audited billing totals. Open-source backbones incurred no API usage charges and used the GPU configurations.
\begin{table}[htb!]
\centering
\caption{\textbf{Approximate average processing time per dialogue.}
Times include tool execution and all assistant responses. Global-to-task results combine both global dialogue subtypes. All values are in seconds.
}
\label{tab:latency}
{\begin{adjustbox}{width=\columnwidth, center}
\begin{tabular}{lccc}
\toprule
Backbone & Single-task & Multi-task & Global-to-task \\
\midrule
Gemini-3-Flash & 29 & 64 & 72 \\
GPT-5 mini & 44 & 72 & 99 \\
Llama 3.3-70B & 45 & 77 & 148 \\
Qwen3-32B & 49 & 52 & 92 \\
Qwen3-8B & 44 & 83 & 99 \\
Qwen3-4B & 36 & 61 & 68 \\
\bottomrule
\end{tabular}
\end{adjustbox}}
\end{table}

\section{\benchmark Dialogue Examples}\label{apd:cxreasondial_examples}
\noindent
Tables~\ref{tab:cxreasondial_dialogue_example}, ~\ref{tab:cxreasondial_dialogue_example_continued}, ~\ref{tab:cxreasondial_cardiomegaly_example}, and ~\ref{tab:cxreasondial_cardiomegaly_example_continued} illustrate user-facing multi-turn interactions in \benchmark.
To avoid redistributing restricted MIMIC-CXR-JPG data, the chest X-ray images and numeric values have been independently replaced with illustrative content. They do not correspond to any individual MIMIC-CXR-JPG study and are shown solely to illustrate the dialogue format.

\begin{table*}[t]
\centering
\small
\caption{\textbf{Illustrative \benchmark dialogue for inspiration assessment (Part I).}
The first user turn includes an illustrative chest X-ray image. The dialogue shows only user-facing turns; all values are illustrative.}
\label{tab:cxreasondial_dialogue_example}
\begin{tabular}{@{}p{0.10\textwidth}>{\raggedright\arraybackslash}p{\dimexpr0.90\textwidth-2\tabcolsep\relax}@{}}
\toprule
\textbf{Role} & \textbf{Dialogue} \\
\midrule
User & How is the degree of lung expansion, or inspiration, evaluated on this chest X-ray, and is it considered adequate?\par\smallskip
\textit{Attached chest X-ray:}\par
{\centering\includegraphics[width=0.25\textwidth]{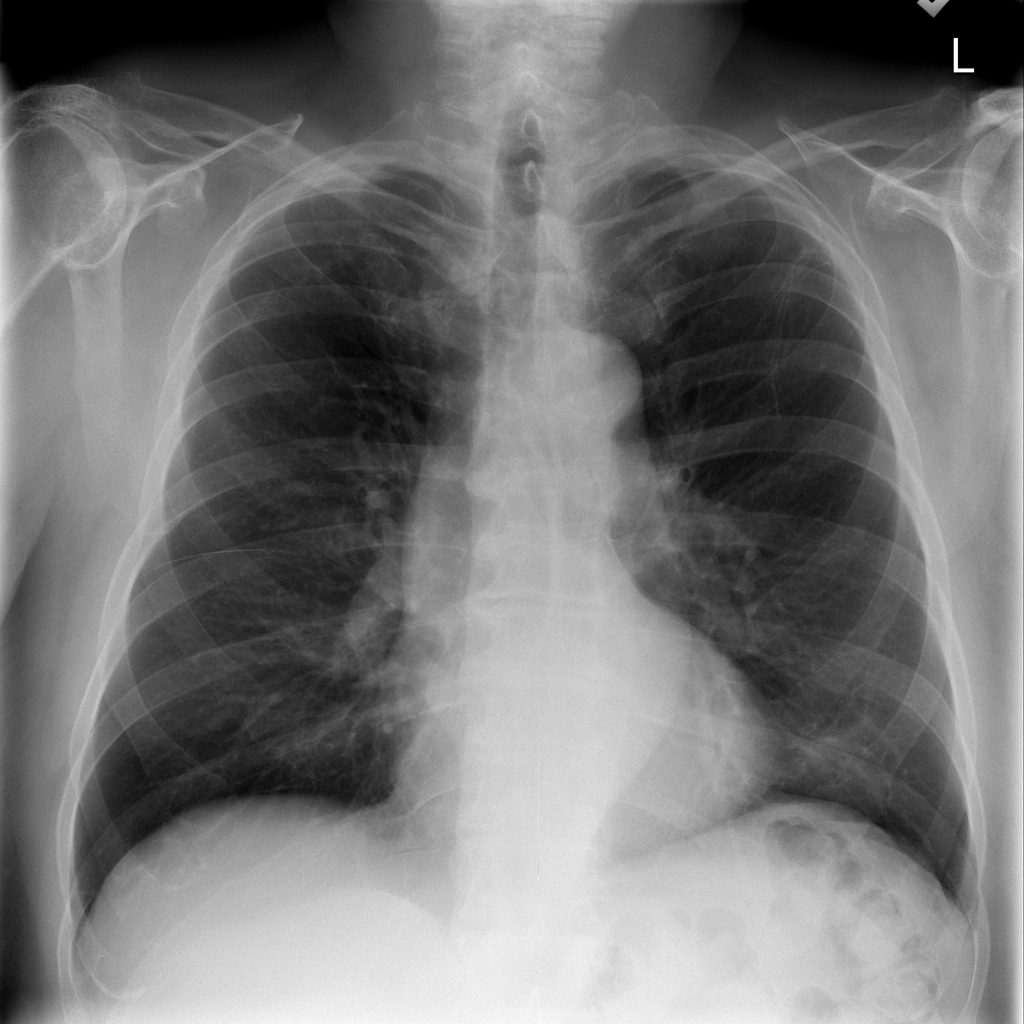}\par}\\
\addlinespace

Agent &
Inspiration is evaluated by counting the right posterior ribs along the mid-clavicular line down to the rib immediately above the right hemidiaphragm. The count is 10, which meets the threshold of 9 or more and therefore indicates adequate inspiration. \\
\addlinespace

User &
Why is adequate inspiration important when interpreting a chest X-ray? \\
\addlinespace

Agent &
Adequate lung expansion improves image quality and helps ensure that thoracic anatomy can be interpreted reliably. \\
\addlinespace

User &
Could you show the landmarks used for this assessment? \\
\addlinespace

Agent &
The annotated image highlights the mid-clavicular line, the right posterior rib immediately above the right hemidiaphragm, and the right hemidiaphragm as the lower boundary for the rib count. \par\smallskip
\textit{Returned visualization:}\par
{\centering\includegraphics[width=0.25\textwidth]{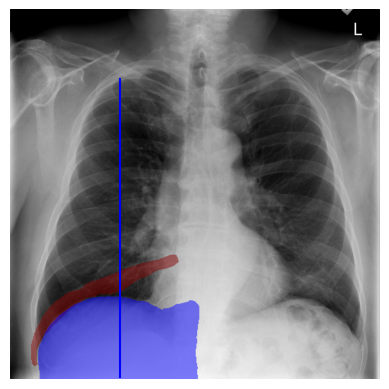}\par}\\

\bottomrule
\end{tabular}
\end{table*}

\begin{table*}[t]
\centering
\small
\caption{\textbf{Illustrative \benchmark dialogue for inspiration assessment (Part II).}
Continuation of Table~\ref{tab:cxreasondial_dialogue_example}.}
\label{tab:cxreasondial_dialogue_example_continued}
\begin{tabular}{@{}p{0.10\textwidth}>{\raggedright\arraybackslash}p{\dimexpr0.90\textwidth-2\tabcolsep\relax}@{}}
\toprule
\textbf{Role} & \textbf{Dialogue} \\
\midrule

User &
Based on the assessment method, what would the conclusion be if the count of right posterior ribs had been lower than 9? \\
\addlinespace

Agent &
If the right posterior rib count along the mid-clavicular line were below 9, the inspiration would be considered inadequate. This threshold helps ensure that image quality is sufficient for reliable interpretation. \\

User &
Can you provide a visual that specifically isolates the right hemidiaphragm as the boundary used for this assessment? \\
\addlinespace

Agent &
The annotated image highlights the right hemidiaphragm, which serves as the lower boundary for identifying the posterior rib used in the inspiration assessment. \par\smallskip
\textit{Returned visualization:}\par
{\centering\includegraphics[width=0.25\textwidth]{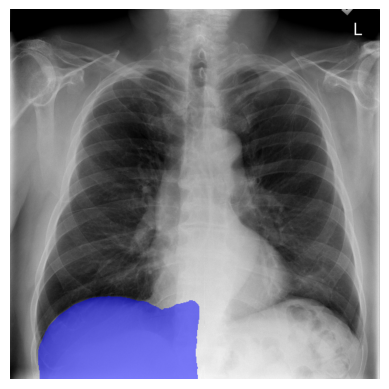}\par}\\
\addlinespace

User &
That explains everything I wanted to know. Thank you for your help. \\
\addlinespace

Agent &
You are welcome. I am glad I could help clarify the assessment. \\
\bottomrule
\end{tabular}
\end{table*}

\begin{table*}[t]
\centering
\small
\caption{\textbf{Illustrative \benchmark dialogue for cardiomegaly assessment (Part I).}
The first user turn includes an illustrative chest X-ray image. The dialogue
shows only user-facing turns; all values are illustrative.}
\label{tab:cxreasondial_cardiomegaly_example}
\begin{tabular}{@{}p{0.10\textwidth}>{\raggedright\arraybackslash}p{\dimexpr0.90\textwidth-2\tabcolsep\relax}@{}}
\toprule
\textbf{Role} & \textbf{Dialogue} \\
\midrule
User & How is the cardiac silhouette assessed on this chest X-ray, and what criteria are used to determine whether it is enlarged?
\par\smallskip
\textit{Attached chest X-ray:}\par
{\centering\includegraphics[width=0.25\textwidth]{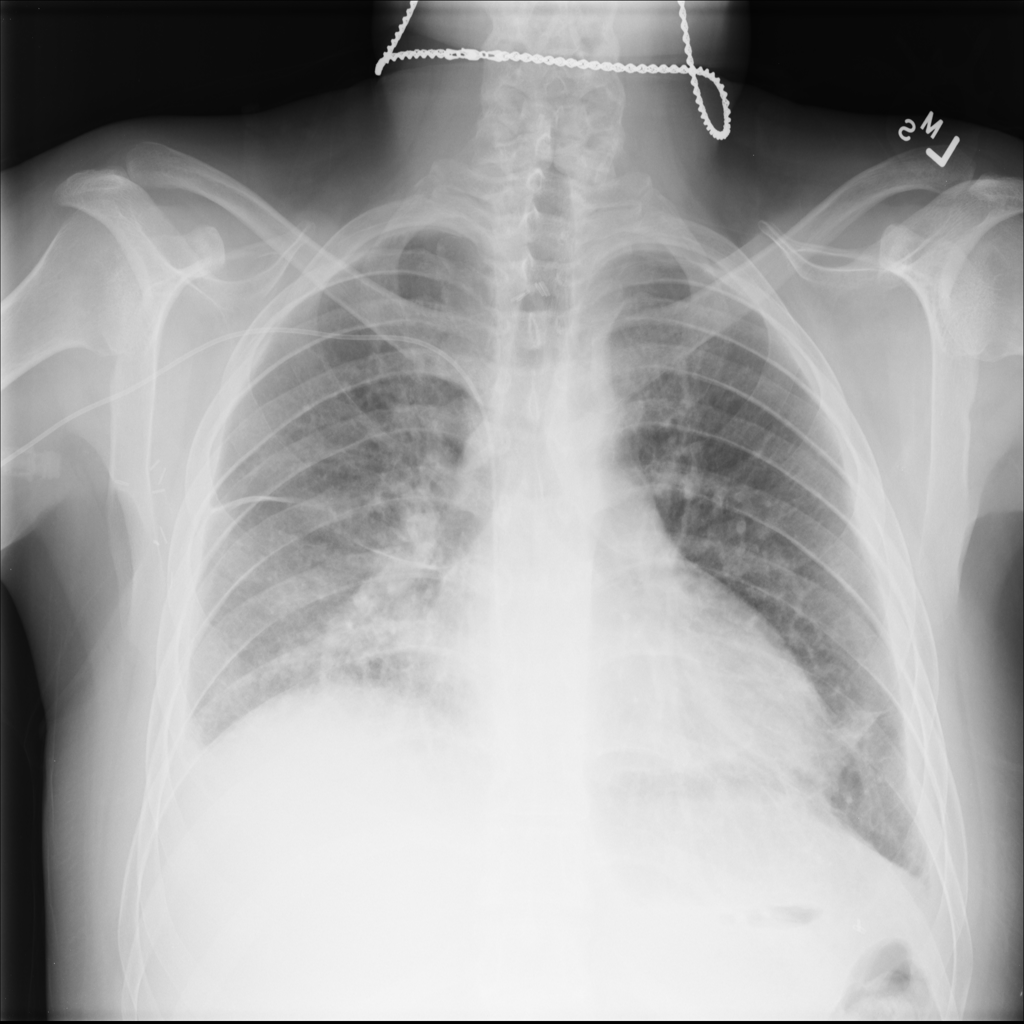}\par}\\
\addlinespace

Agent &
Cardiac size is assessed using the cardiothoracic ratio (CTR), calculated as the maximal cardiac width divided by the maximal thoracic width. Here, the cardiac width is 1320 pixels and the thoracic width is 2400 pixels, yielding a CTR of 0.55. This meets the enlargement threshold for both PA and AP projections; therefore, the cardiac silhouette is enlarged, consistent with cardiomegaly. \\
\addlinespace

User &
Could you show how the cardiac and thoracic widths were measured on the image? \\
\addlinespace

Agent &
The annotated image marks the maximal cardiac width with a red line and the maximal thoracic width with a blue line. These measurements are used to calculate the CTR of 0.55. \par\smallskip
\textit{Returned visualization:}\par
{\centering\includegraphics[width=0.25\textwidth]{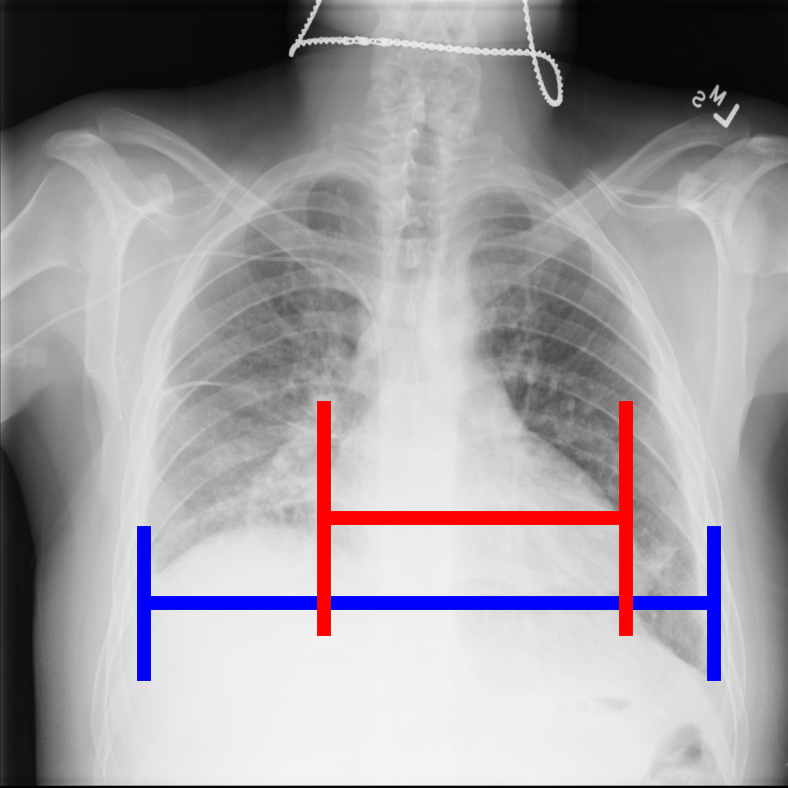}\par}\\
\addlinespace

User &
Why is a CTR of 0.55 considered enlarged regardless of whether the image is PA or AP? \\
\addlinespace

Agent &
The threshold is 0.50 for PA images and 0.55 for AP images. A CTR of 0.55 meets or exceeds both thresholds, so it is classified as enlarged for either projection. \\
\addlinespace

User &
How would the interpretation change if the CTR were between 0.50 and 0.55? \\
\addlinespace

Agent &
In that range, the interpretation depends on projection. The cardiac silhouette would be enlarged on a PA image because the PA threshold is 0.50, but normal on an AP image because the AP threshold is 0.55. \\
\bottomrule
\end{tabular}
\end{table*}

\begin{table*}[t]
\centering
\small
\caption{\textbf{Illustrative \benchmark dialogue for cardiomegaly assessment (Part II).}
Continuation of Table~\ref{tab:cxreasondial_cardiomegaly_example}.}
\label{tab:cxreasondial_cardiomegaly_example_continued}
\begin{tabular}{@{}p{0.10\textwidth}>{\raggedright\arraybackslash}p{\dimexpr0.90\textwidth-2\tabcolsep\relax}@{}}
\toprule
\textbf{Role} & \textbf{Dialogue} \\
\midrule
User &
What would a CTR of 0.45 indicate? \\
\addlinespace

Agent &
A CTR of 0.45 is below both the PA and AP thresholds and would therefore indicate a normal cardiac silhouette. \\
\addlinespace

User &
If the projection were unknown and the CTR were in the borderline range, how should the finding be reported? \\
\addlinespace

Agent &
The finding should be reported as borderline, with the projection identified before assigning a final classification. A CTR between 0.50 and 0.55 is interpreted differently for PA and AP images. \\
\addlinespace

User &
Could you also show the lung boundaries used to obtain the thoracic width? \par\smallskip
\textit{Returned visualization:}\par
{\centering\includegraphics[width=0.25\textwidth]{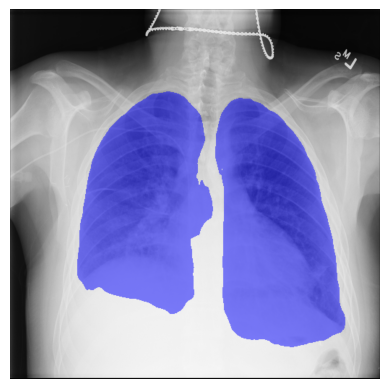}\par}\\
\addlinespace

Agent &
The annotated image highlights the bilateral lung boundaries in blue, indicating the regions used to measure the maximal thoracic width. \\
\addlinespace

User &
Thank you. That clarifies the assessment. \\
\addlinespace

Agent &
You are welcome. \\
\bottomrule
\end{tabular}
\end{table*}

\end{document}